\documentclass[journal]{IEEEtran}
\usepackage{graphics}
\usepackage{graphicx}
\usepackage{subfigure}
\usepackage{color,soul,tcolorbox}
\usepackage{amssymb}
\usepackage{amsfonts}
\usepackage{amsmath}
\usepackage{upgreek}
\usepackage{dsfont}
\usepackage{floatflt}
\usepackage{setspace}
\usepackage{enumitem}
\usepackage{mathtools}
\usepackage{tabularx}
\usepackage{url}
\usepackage{cite}
\usepackage[lined,ruled,commentsnumbered]{algorithm2e}
\newcommand{\overbar}[1]{\mkern 1.5mu\overline{\mkern-1.5mu#1\mkern-1.5mu}\mkern 1.5mu}
\newenvironment{bluenote}{\par\color{black}}{\par}
\newcolumntype{L}[1]{>{\raggedright\arraybackslash}p{#1}}
\newcolumntype{C}[1]{>{\centering\arraybackslash}p{#1}}
\newcolumntype{R}[1]{>{\raggedleft\arraybackslash}p{#1}}

\begin{document}
\title{Novel Subtypes of Pulmonary Emphysema \\ Based on Spatially-Informed Lung Texture Learning}

\author{Jie Yang, Elsa D. Angelini, Pallavi P. Balte, Eric A. Hoffman, \\ John H.M. Austin, Benjamin M. Smith, R. Graham Barr, and Andrew F. Laine*
\thanks{This work was supported by NIH/NHLBI R01-HL121270, R01-HL077612, RC1-HL100543, R01-HL093081 and N01-HC095159 through N01-HC-95169, UL1-RR-024156 and UL1-RR-025005. \textit{Asterisk indicates corresponding author}.}
\thanks{Jie Yang and Andrew F. Laine are with the Department of Biomedical Engineering, Columbia University, New York, NY, USA (e-mail: jy2666@columbia.edu; al418@columbia.edu).}
\thanks{Elsa D. Angelini is with the Department of Biomedical Engineering, Columbia University, New York, NY, USA, and the NIHR Imperial BRC, ITMAT Data Science Group, Department of Metabolism-Digestion-Reproduction, Imperial College, London, UK (e-mail: e.angelini@imperial.ac.uk).}
\thanks{Pallavi P. Balte is with the Department of Medicine, Columbia University Medical Center, New York, NY, USA.}
\thanks{Eric A. Hoffman is with the Departments of Radiology, Medicine and Biomedical Engineering, University of Iowa, Iowa City, IA, USA.}
\thanks{John H.M. Austin is with the Department of Radiology, Columbia University Medical Center, New York, NY, USA.}
\thanks{Benjamin M. Smith is with the Department of Medicine, Columbia University Medical Center, New York, NY, USA, and the Department of Medicine, McGill University Health Center, Montreal, QC, Canada.}
\thanks{R. Graham Barr is with the Department of Medicine and Epidemiology, Columbia University Medical Center, New York, NY, USA.}}


\maketitle
\begin{abstract}
	
Pulmonary emphysema overlaps considerably with chronic obstructive pulmonary disease (COPD), and is traditionally subcategorized into three subtypes previously identified on autopsy. Unsupervised learning of emphysema subtypes on computed tomography (CT) opens the way to new definitions of emphysema subtypes and eliminates the need of thorough manual labeling. However, CT-based emphysema subtypes have been limited to texture-based patterns without considering spatial \textcolor{black}{location}. In this work, we introduce a standardized spatial mapping of the lung for quantitative study of lung texture \textcolor{black}{location}, and propose a novel framework for combining spatial and texture information to discover spatially-informed lung texture patterns (sLTPs) that represent novel emphysema subtypes. Exploiting two cohorts of full-lung CT scans from the MESA COPD  and EMCAP studies, we first show that our spatial mapping enables population-wide study of emphysema spatial location. We then evaluate the characteristics of the sLTPs discovered on MESA COPD, and show that they are reproducible, able to encode standard emphysema subtypes, and associated with physiological symptoms.

\end{abstract}
 

\IEEEpeerreviewmaketitle

\section{Introduction}

Pulmonary emphysema is morphologically defined  by the enlargement of airspaces with destruction of alveolar walls distal to the terminal bronchioles \cite{emphysema}. Emphysema overlaps considerably with chronic obstructive pulmonary disease (COPD), which is currently the fourth leading cause of death in the world, and is projected to be the third leading cause of death in 2020 \cite{gold}.
Based on small autopsy series, pulmonary emphysema is traditionally subcategorized into three standard subtypes, which can be visually assessed on computed tomography (CT) of the lung, using the following definitions: 

\begin{itemize}[leftmargin=0.35cm,labelwidth=0.1cm]
\item \textit{Centrilobular emphysema} (CLE): low-attenuation regions surrounded by normal lung, and located centrally in the secondary pulmonary lobules \cite{subtypes}. Classically, its distribution is predominantly in the apical regions of the lungs; 
\item \textit{Panlobular emphysema} (PLE): low-attenuation regions which are uniformly diffuse in the secondary pulmonary lobules \cite{ben}. Classically, its distribution is predominantly in the basal regions of the lungs; 
\item \textit{Paraseptal emphysema} (PSE): low-attenuation regions adjacent to pleura and to intact interlobular septa, typically found in juxtapleural lobules adjacent to mediastinal and costal pleura \cite{subtypes}. Classically, its distribution is predominantly in the upper and middle lung zones.
\end{itemize}

The three standard emphysema subtypes are associated with distinct risk factors and clinical manifestations \cite{riskfactor1}, and are likely to represent different diseases. However, given that these subtypes were initially defined at autopsy before the availability of CT scanning, there have been disagreements among pathologists on the very existence of such pure subtypes \cite{Anderson64}, and a large emphysema study on 1,800 autopsies in \cite{Auerbach72} ignored them completely, mainly for practical reasons. Radiologists' interpretation of these subtypes on CT scans is labor-intensive, with substantial intra- and inter-rater variability \cite{ben}.

Automated CT-based analysis enables \textit{in vivo} study of emphysema patterns, and has received increasing interest recently \cite{met,review_lungtexture}, either via supervised learning for replicating emphysema subtype labeling as in \cite{lbp2_tmi10, rad_2012, texton_miccai2010,texton_spie14, ross2017bayesian}, or via unsupervised learning for the discovery of new emphysema subtypes as in \cite{ltp_mit16,ltp_isbi15,ltp_mcv16}. 

Preliminary CT-based clinical studies suggest that regional analysis will be instrumental in advancing the understanding of multiple pulmonary diseases \cite{spatial_mp12}. In the case of emphysema, it is suspected that different emphysema subtypes affect the lungs in \textcolor{black}{preferred anatomical regions}. But \textcolor{black}{physiological  understanding} of how many subtypes exist, how they evolve in time and how they vary with spatial location is still unsolved. To date, categorization of emphysema on CT images has relied only on analysis of local textural patterns, using either grey-level co-occurrence matrix (GLCM) features \cite{rad_2012,ltp_mit16},  texton features  \cite{texton_miccai2010,texton_spie14}, or local binary pattern (LBP) features \cite{lbp2_tmi10}. All these approaches use intensity information without consideration of spatial \textcolor{black}{location}. 

\begin{figure*}[t]
	\centering
	\includegraphics[width=.99\linewidth]{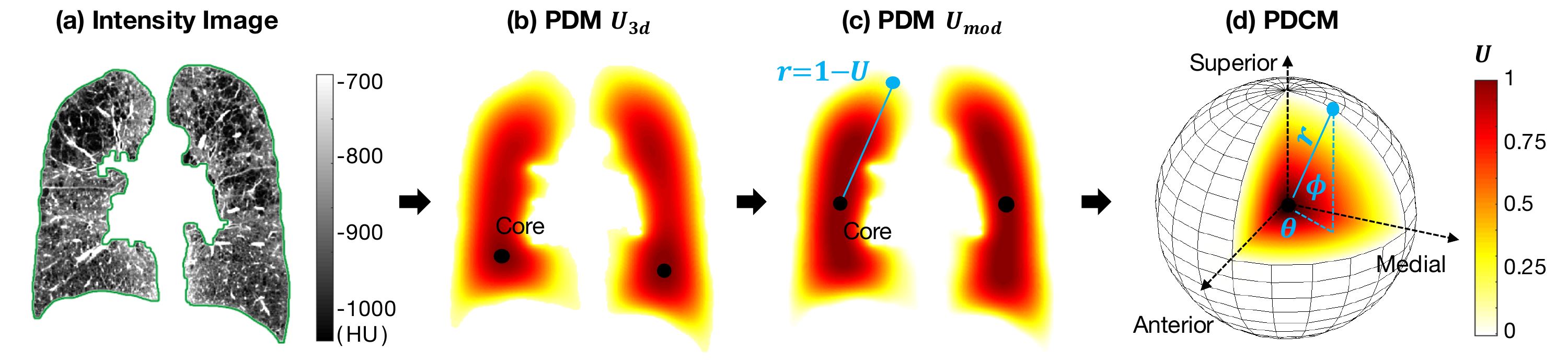}
	\caption{Illustration of the lung shape spatial mapping: (a) Original intensity image (visualized on a coronal slice, with the green contour indicating the boundary of lung mask); (b) Corresponding Poisson distance map (PDM) $U_{3d}$ with values in range $[0,1]$ that measure the ``peel to core" 3D distance to the lung mask external surface; (c) Modified PDM $U_{mod}$ for comparable core locations between subjects; (d) 3D conformal mapping of the lung PDM to a sphere leading to a Poisson distance conformal map (PDCM) where pixels are assigned three coordinate values $(r,\theta,\phi)$ which enable to distinguish superior vs. inferior, anterior vs. posterior and medial vs. lateral positions, in addition to ``peel to core" distance.} 
	\label{Fig:illustration_pdcm}
\end{figure*}

In two previous studies \cite{ltp_isbi15,ltp_mcv16}, we proposed to use local textural patterns to generate unsupervised lung texture patterns (LTPs) followed by LTP-grouping based on their spatial co-occurrence in local neighborhoods. Such separate use of intensity and spatial information cannot guarantee spatial and textural \textcolor{black}{homogeneity} of the final LTPs. 

In this study, we propose to perform  discovery of LTPs via unsupervised clustering of joint spatial and textural information of local texture patterns. Spatial information can be inferred from crude partitioning of the lung with subdivisions of Cartesian coordinates or by segmenting the lung into zones (e.g. upper, lower) \cite{ben} or lobes \cite{lobe}. However, such approaches have limited spatial precision and lack relative information such as peripheral versus central positioning, which is important in defining paraseptal emphysema and subpleural bullae. 

We introduced in \cite{sltp_miccai16} a new standardized lung shape spatial mapping, called Poisson distance conformal mapping (PDCM), which enables detailed, precise and standardized mapping of voxel positions with respect to the lung surfaces. 
This paper further refines the PDCM algorithm and exploits it for the study of emphysema spatial patterns across populations of CLE-, PLE- and PSE-predominant subjects, \textcolor{black}{without registration being required further than orientation alignment}. 
This paper also provides an exhaustive description of the framework for combining spatial and texture information in the unsupervised discovery of \emph{emphysema-specific} texture patterns, which are called spatially-informed LTPs (sLTPs).
Exploiting a cohort of 317 full-lung CT scans from the MESA COPD study \cite{ben}, and 22 longitudinal CT scans from the EMCAP study \cite{emcap}, the discovered sLTPs are extensively evaluated in terms of reproducibility with respect to training sets, labeling task and scanner generations, ability to encode standard emphysema subtypes, and associations with respiratory symptoms.

\section{Method}\label{section_method}

\subsection{\textcolor{black}{Overview}}

\begin{bluenote}
The proposed framework is structured in four main steps to model the spatial and texture features within emphysema-like lung, and generate the emphysema-specific sLTPs:
\begin{enumerate}[leftmargin=0.45cm,labelwidth=0.1cm]
	\item Generate spatial mapping of the lung masks: mapping voxels within the lung masks into a custom Poisson distance map (PDM) to encode the ``peel to core" distance, and a conformal mapping to distinguish superior versus inferior, anterior versus posterior and medial versus lateral voxel positions;   
	\item Encode regions of interest (ROIs) within emphysema-like lung: sampling ROIs from emphysema segmentation masks, and generating spatial features (based on spatial mapping) and texture features of each ROI;
	\item Discover an initial set of LTPs: clustering training ROIs into a large number of clusters, based on texture features, and then iteratively augment the LTPs with spatial information via regularization;
	\item  Generate the final set of sLTPs: measure the similarity between LTPs in the initial set, group similar / redundant LTPs and generate the final set of sLTPs via partitioning the similarity graph.
\end{enumerate}
\end{bluenote}

We now detail these three steps individually.

\subsection{Spatial Mapping of the Lung Masks}\label{section_method_pdcm}

To generate spatial mapping of the lung masks, we first use the concept of Poisson distance map (PDM), introduced in \cite{poisson}, to encode the shape of individual lung masks $V$. PDM is commonly used for characterizing the silhouette of an object via continuous labeling of voxel positions with scalar field values $U$ in the range of $[0,1]$. In our case, the field value $U$ encodes the ``peel to core" distance between a given voxel and the external lung surface $\partial V$. This field is computed by solving the following Poisson equation:
\begin{equation}\label{poisson}
\begin{aligned}
{}&\Delta U(x,y,z)=-1, \ \mathrm{for} \ (x,y,z)\in V \\
& \mathrm{subject \ to} \ U(x,y,z)=0, \ \mathrm{for} \ (x,y,z)\in \partial V
\end{aligned}
\end{equation}
\noindent where $\Delta U = U_{xx}+U_{yy}+U_{zz}$. 

The solution for $U$ proposed in \cite{poisson} is guaranteed to be smooth according to \cite{poisson_tmi}. It has the advantage of generating distance values that are sensitive to global shape characteristics, unlike other distance metrics (e.g. Euclidian or Metropolis distances) which exploit single contour points. PDM can therefore reflect rich shape properties of the lung.

The core of the PDM is the set of voxels (one or very few) with the largest $U$ value. 
The PDM generated from a lung surface generally exhibits nice star-shaped profiles when viewed in axial cuts, with a unique maxima in the center. On the other hand, core positions can vary greatly among subjects along superior-inferior axis, due to variable morphologies of the lungs, especially near the heart and at the base. We illustrate an example in Fig. \ref{Fig:illustration_pdcm} (b) where the PDM generated with Equation (\ref{poisson}) has core point(s) located very low within the lung rather than concentrated toward the middle of the longitudinal axis. 
We propose the following approach to calibrate lung PDMs targeting high values of $U$ concentrated near the skeleton of the lung shapes and in the mid-level slices. 

We denote $U^{max}(S_i)$ the maximal in-slice value of $U$, where $S_i$ is the axial slice level \textcolor{black}{with $i$ in ascending order from the apex}. We denote $S_{{V\%}}$ the slice level with $V\%$ of total lung volume above.
A \textcolor{black}{normalized version (denoted as $U_{2d}$}), of the \textcolor{black}{original PDM (denoted as $U_{3d}$}), is then defined, per axial slice $S_i$, as $U_{2d}(S_i) = U_{3d}(S_i)/U_{3d}^{max}(S_i)$.  

We further modify $U$ by combining $U_{3d}$ and $U_{2d}$ values. 
First, two axial slice levels $S_{i_u'}$ and $S_{i_d'}$, corresponding to the most apical and  basal slice levels of local maxima in $U_{3d}$, are identified as:
\textcolor{black}{\begin{equation}
\begin{aligned}\label{kud}
{}& i_u' = \underset{x}{\mathrm{argmax}} \big[U_{3d}^{max}(S_i)<U_{3d}^{max}(S_{x}),\forall \ i<x \big] \\
  & i_d' = \underset{x}{\mathrm{argmin}} \big[U_{3d}^{max}(S_i)<U_{3d}^{max}(S_{x}),\forall \ i>x \big] \\
\end{aligned}
\end{equation}}
\noindent We then define two reference slice levels $S_{i_u}$ and $S_{i_d}$ as:
\begin{equation}\label{2575}
S_{i_u}=\min(S_{{25\%}},S_{i_u'}) \mathrm{\ and \ } S_{i_d}=\max(S_{{75\%}},S_{i_d'})
\end{equation}
\noindent The reference levels $S_{i_u}$ and $S_{i_d}$ are exploited to ensure that the modified core regions reach at least extremal levels $S_{{25\%}}$ and $S_{{75\%}}$, with the following modification of the $U$ values into the \textcolor{black}{modified PDM (denoted as $U_{mod}$)}:
\begin{equation}
\begin{aligned}\label{mod}
{}& U_{mod} (S_i)=U_{2d} (S_i), \ \forall  \ i_u\leqslant i\leqslant i_d \\ 
& U_{mod} (S_i)=U_{3d} (S_i)/U_{3d}^{max}(S_{i_u}), \ \forall  \ i<i_u \\
& U_{mod} (S_i)=U_{3d} (S_i)/U_{3d}^{max}(S_{i_d}), \ \forall  \ i>i_d
\end{aligned}
\end{equation}

We illustrate in Fig. \ref{Fig:illustration_pdcm} (c) an example of $U_{mod}$ which takes similar maximal values (equal to 1) over a large mid-level extent along the superior-inferior axis and exhibits decreasing values when moving toward the apex or the base of the lung. 

This simple calibration enables us to equip the PDM with a coordinate system centered at a core localized on axial slice level $S_{{50\%}}$ (ensuring a balanced numbers of voxels above and below), where the core is defined as the point with $U_{mod}=1$, and closest to the 2D center of mass, for the sake of simplicity.

To uniquely encode 3D voxel positions, we define radial values $r=1-U_{mod}$ and add conformal mapping of voxels positions onto a sphere, generating a Poisson distance conformal map (PDCM). We encode superior versus inferior, anterior versus posterior and medial versus lateral voxel positioning via latitude and longitude angles $(\theta,\phi)$ with respect to the PDM core defined above and standard image axis. The generation of the spatial PDCM mapping is illustrated in Fig. \ref{Fig:illustration_pdcm} (d).

The PDCM spatial mapping will be exploited for sLTP learning, and also to study population-based spatial \textcolor{black}{location} of emphysema, as reported in Section \ref{section_result_pdcm}.

\subsection{Texture and Spatial Features}\label{section_method_feature}

\subsubsection{\textcolor{black}{Prior Emphysema Segmentation and ROI Sampling}}
Texture and spatial analysis is performed within local ROIs centered on a subset of lung voxels. 
Sampling ROIs from emphysema-like lung requires prior emphysema segmentation. In this study, we exploited a training cohort of full-lung CT scans and their associated emphysema masks, which are generated using both a thresholding-based voxel selection and a hidden Markov measure field (HMMF) segmentation \cite{hmmf_tmi14}. For thresholding, voxels with attenuation below $-950$ HU are selected. This threshold has been previously validated against autopsy specimens and is commonly used in large clinical studies \cite{hmmf_miccai16}. The HMMF segmentation enforces spatial coherence of the labeled emphysematous regions, and relies on parametric modeling of intensity distributions within emphysematous and normal lung tissues to adapt to individual and scanner variability. With the two sets of emphysema masks, percent emphysema measures quantify the proportion of emphysematous voxels within the lung region, and are denoted $\%emph_{-950}$ and $\%emph_{\mathrm{HMMF}}$.

We experimented several options for ROI sampling in preliminary implementations such as keypoint sampling in \cite{ltp_isbi15} and regular sampling in \cite{ltp_mcv16}. In this study, we use the systematic uniform random sampling (SURS) strategy as suggested in \cite{surs} for use on lung CT scans. Each individual lung mask is randomly sampled via dividing the bounding box of the lung into 3D stacks, and then selecting voxels per stack with a random shift of positions. Two parameters are used for the sampling: $\beta_1$ is used for the random shift of positions and $\beta_2$ is used to set the number of sampled voxels per stack. The SURS sampling ensures even representation of all lung regions while introducing variability in the position of sampled points with the random shift parameter $\beta_1$. Only ROIs with both percent emphysema $\%emph_{-950}>1\%$ and $\%emph_{\mathrm{HMMF}}>1\%$ are retained for training to ensure sufficient representation of emphysematous regions (i.e. each training ROI has a minimal proportion of emphysema but can be a mixture of normal and emphysematous tissues).

\subsubsection{\textcolor{black}{Texture Features}}\label{section_texture_feature}

We use texton-based texture feature to characterize each ROI, which models texture as the repetition of a few basic primitives (called textons), and was shown to outperform other texture features in unsupervised lung texture learning in \cite{ltp_mcv16}. A texton codebook is constructed by retaining the cluster centers (textons) of intensity values from small-sized training patches. The clustering is performed with $K$-means. By projecting all small-sized patches of a ROI onto the codebook, the texton-based feature of the ROI is the normalized histogram of texton frequencies. 

\subsubsection{\textcolor{black}{Spatial Features}}\label{section_spatial_feature}

To generate spatial features of individual ROIs, we divide the lung masks into lung sub-regions via discretizing our lung shape spatial mapping. For the sake of simplicity, we define lung sub-regions by dividing $r \in [0, 1]$ into 3 regular intervals to distinguish core to peel regions, dividing $\theta \in [0, 2\pi]$ into 4 regular intervals to distinguish anterior, medial, posterior and lateral regions, and dividing $\phi \in [-\pi/2, \pi/2]$ into 3 regular intervals to distinguish inferior, mid-level and superior regions.  The spatial feature of each ROI is a one-hot vector indicating the lung sub-region it belongs to. Ordering of the bins that represent the sub-regions is done via arbitrary spatial rastering as no assumption needs to be made on spatial adjacency of adjacent bins. 

\subsection{Initial Augmented LTPs}\label{section_method_ltp}

\textcolor{black}{Our discovery of spatially-informed lung texture patterns (sLTPs) is formulated as an unsupervised clustering problem. One key factor in unsupervised clustering is the choice of number of clusters. The algorithm is expected to find finer-grained emphysema types than the three standard subtypes. Therefore, the number of clusters should be large enough to handle the diversity of textures encountered in the lung volumes (i.e. good intra-cluster homogeneity), and on the other hand, be small enough to avoid redundancy (i.e. good inter-cluster differences) for better clinical interpretation. A simple one-stage clustering is suboptimal since it requires tuning or a pre-fixed number of clusters, and may not be able to preserve rare patterns. We propose a two-stage learning strategy, where we first generate an empirically large number of fine-grained lung texture patterns (LTPs), and then group similar LTPs to produce the final set of sLTPs, according to a dedicated metric.}

LTPs $\{LTP_k\}$ ($\{\cdot\}$ denotes a set of variables hereafter) are characterized by their spatial and texture feature centroids, which are encoded as histograms, and are enforced for intra-class similarity and inter-class separation. For a given $LTP_k$, its \textit{texture} centroid $\overbar{FT}_{LTP_k}$ and \textit{spatial} centroid $\overbar{FS}_{LTP_k}$ are computed as:
\begin{equation}\label{FTS}
	\Big[\overbar{FT}_{LTP_k}, \ \overbar{FS}_{LTP_k}\Big] = \frac{1}{|\Lambda_{LTP_k}|} \sum_{x\in \Lambda_{LTP_k}}\Big[FT_x, \ FS_x\Big]
\end{equation}
where $FT_x$ and $FS_x$ are respectively the texture feature and spatial feature of a ROI $x$, and $\Lambda_{LTP_k}$ denotes the set of ROIs that are labeled as $LTP_k$. 

An initial set of LTPs is generated by clustering with \textit{texture} features, and is then augmented with \textit{spatial} regularizations via iteratively updating $\{\overbar{FT}_{LTP_k}$, $\overbar{FS}_{LTP_k}\}$ and $\{\Lambda_{LTP_k}\}$. The generation and augmentation of LTPs are summarized in Algorithm \ref{algo1}. 

Designing proper distance metrics for histograms plays a crucial role in many computer vision tasks. Two popular choices are the $\chi^2$ and the $\ell^2$ distance metrics. The latter equally weights distances of all bins and is favored to compare one-hot vectors, while the former is a weighted distance and is favored to compare probability distributions. In our case, texture feature histograms encode distributions over textons, and the $\chi^2$ metric is used. On the other hand, spatial features are sparse one-hot vectors for individual ROIs and we chose the $\ell^2$ metric to favor spatial centroids being concentrated in specific lung sub-regions. We therefore propose a mixed  \raisebox{1pt}{$\chi^2$}-$\ell^2$ similarity metric to enforce spatial concentration of LTPs while preserving their intra-class textural homogeneity: 
\textcolor{black}{\begin{align}\label{iter}
	& \big\{\Lambda_{LTP_k}^{(t)}{}\big\}^*_{\{\lambda, W, \gamma \}} = 
	\underset{\{\Lambda_{LTP_k}^{(t)}\}}{\mathrm{argmin}}\sum_{k}\sum_{x \in \Lambda_{LTP_k}^{(t)}} \\ 
	& \displaystyle \chi^2 \big(FT_{x},\overbar{FT}_{LTP_k}^{(t-1)}\big)
	+ \lambda \cdot W \cdot \left|\left|FS_{x}-\overbar{FS}_{LTP_k}^{(t-1)}\right|\right|_2^2  + \nonumber \\
	& \displaystyle \gamma \cdot \mathds{1}\Big[ \chi^2 \big(FT_{x},\overbar{FT}_{LTP_k}^{(t-1)}\big) 
	> \max_{x' \in \Lambda_{LTP_k}^{(t-1)}}\chi^2 \big(FT_{x'},\overbar{FT}_{LTP_k}^{(t-1)}\big)\Big] \nonumber 
\end{align}}
\begin{bluenote}
\noindent where $\big\{\Lambda_{LTP_k}^{(t)}\big\}^*_{\{\lambda, W, \gamma \}}$ denotes the optimal value identified with a set of parameters  $\{\lambda, W, \gamma \}$ at iteration $t$. 
The first distance metric $\chi^2(\cdot)$ measures the $\chi^2$ distance between the textural feature of a ROI $x$ and $LTP_k$. 
The second distance metric $||\cdot||_2^2$ measures the $\ell^2$ distance between the spatial feature of a ROI $x$ and $LTP_k$. 
A textural penalty term is then introduce as the third term, where $\mathds{1}$ is the indicator function.

Minimization of Equation (\ref{iter}) (step 1 in Algorithm \ref{algo1}) is performed via exhaustive search over all possible values of $\{\Lambda_{LTP_k}^{(t)}\}$. Update of LTP centroids (step 2 in Algorithm \ref{algo1}) is performed after relabeling each ROI to the LTP to which it has the smallest weighted feature distances without turning on the penalty.
\end{bluenote}

\setlength{\textfloatsep}{12pt}
\begin{algorithm}[t]
	\small{
	\setstretch{1.275}
	\SetKwData{Left}{left}\SetKwData{This}{this}\SetKwData{Up}{up}
	\SetKwFunction{Union}{Union}\SetKwFunction{FindCompress}{FindCompress}
	\SetKwInOut{Input}{Input}\SetKwInOut{Output}{Output}
	 \Input{$N_{LTP}$ : Target number of LTPs;
	 \\ $\{x,FT_x,FS_x \}$ : Training ROIs $x$ along with their \\
	texture features $FT_x$ and spatial features $FS_x$.
	 }
	\Output{$\{\overbar{FT}_{LTP_k}$, $\overbar{FS}_{LTP_k}\}_{k=1,...,N_{LTP}}$ : LTP texture and spatial feature centroids.
	 }
	\textbf{Procedure:}\\
	- Cluster training ROIs $\{x\}$ into $N_{LTP}$ clusters with $\{FT_{x}\}$, using $K$-means.\\
	- \textcolor{black}{Set t = 0}, and initialize $\Lambda_{LTP_k}^{(0)}$ ($k=1,...,N_{LTP}$) with the $N_{LTP}$ LTPs.\\
	- For each $k$, compute $\overbar{FT}_{LTP_k}^{(0)}$, $\overbar{FS}_{LTP_k}^{(0)}$ based on $\Lambda_{LTP_k}^{(0)}$. \\
	\textcolor{black}{
	\setstretch{1.5}
	\While{$t = 0$ $\textnormal{or}$ $\{\Lambda_{LTP_k}^{(t)}\}\neq \{\Lambda_{LTP_k}^{(t-1)}\}$}	
	{ 	
		1. $t = t+1$;\\
		2. $\{\Lambda_{LTP_k}^{(t)}\} \leftarrow \{\Lambda_{LTP_k}^{(t)}\}^*$  following Equation (\ref{iter});\\
		3. Compute $\{\overbar{FT}_{LTP_k}^{(t)}$, $\overbar{FS}_{LTP_k}^{(t)}\}$ based on $\{\Lambda_{LTP_k}^{(t)}\}$.
		}
	}}	
	\caption{Generating and Augmenting LTPs}\label{algo1}
\end{algorithm}

\noindent\textbf{Parameter $W$}: This parameter is used to scale contributions between \textcolor{black}{textural distance and spatial distance} terms so that $\lambda$ can be tuned within a small range of values. We defined it as:
\begin{equation}\label{W}
	W=\frac{SST_{T}}{SST_{S}}=\frac{\sum\nolimits_x\chi^2 \left(FT_{x},\sum_xFT_x/N \right)}
	{\sum\nolimits_x\left|\left|FS_{x}-\sum_xFS_x/N\right|\right|_2^2}
\end{equation}
where $SST_{T}$ and $SST_{S}$ are respectively the texture and spatial \textit{total} sum-of-square distances, computed on the whole $N$ training ROIs to measure the overall diversity of texture and spatial features.

\noindent \textbf{Parameter $\lambda$}: This parameter controls the spatial regularization which will inevitably decrease textural homogeneity of individual LTPs. The value of $\lambda$ is set as follows. First we define
$SSW_{T}$ as the initial sum-of-square \textit{within-cluster} homogeneity of texture features without spatial regularization:
\textcolor{black}{\begin{equation}
SSW_{T} = \sum\nolimits_k\sum\nolimits_{x\in \Lambda_{LTP_k}^{(0)}}\chi^2\left(FT_{x},\overbar{FT}_{LTP_k}^{(0)}\right)
\end{equation}}
\noindent Then we define $SSW_{T}^{\lambda}$ as the $SSW_T$ measured on augmented LTPs with spatial regularization enforced with $\lambda \in [0,2]$. Final value of $\lambda$ is set to:
\begin{equation}\label{tune-lambda}
\begin{aligned}
{}& \lambda^* = \underset{\lambda}{\mathrm{argmax}}\big[\Delta SSW_{T}(\lambda) < L_{T}\big] \\
& \mathrm{where \ } \Delta SSW_{T}(\lambda)= \frac{SSW_{T}^{\lambda}-SSW_{T}}{SSW_{T}}\%
\end{aligned}
\end{equation} 
\noindent In the context of unsupervised discovery, we hereby spatially regularize the augmented LTPs via an empirically acceptable textural homogeneity loss with the threshold $L_{T}$ (set based on data observations, as reported in Section \ref{section_result}).

\noindent \textbf{Parameter $\gamma$}:
This parameter weights the textural penalty term which is used for ROI labeling. 
We set $\gamma=\infty$ to prevent a ROI from being labeled to a spatially preferred but texturally dissimilar LTP.

\subsection{Final Spatially-Informed LTPs (sLTPs)}\label{section_method_sltp}

In this final step, we generate sLTPs by partitioning a weighted undirected graph $G$ where nodes are the $N_{LTP}$ initial augmented LTPs. To define weighted edges between nodes, we rely on replacement tests. We first define $N_{LTP}$ subsets of augmented LTPs as  $\{ LTP_k\}_{k \ne i}$ (i.e. without $LTP_i$ in the subset of LTPs) for $i=1,2,...,N_{LTP}$. Labeling again all \textcolor{black}{training} ROIs with these subsets, we defined $N_{LTP}$ sets of labeled data $\Lambda_{LTP_{i\rightarrow j}}$ as the ROIs labeled as $LTP_j$ when using $\{ LTP_k\}_{k \ne i}$. In the replacement tests, a ROI with a textural distance to $LTP_k$ exceeding the maximal within-cluster textural distance of $LTP_k$ is not re-labeled. Therefore, defining $N_{i\rightarrow j} =| \Lambda_{LTP_{i\rightarrow j}}| $, we guarantee that $\sum_k{N_{i\rightarrow k}}/N_{i}\leqslant 1$ for $N_i=|\Lambda_{LTP_i}|$ when all augmented LTPs are used for labeling.
We define similarity weights $G_{i,j}$ as a measure of replacement ratios of $LTP_i$ into $LTP_j$ and vice versa:
\begin{equation}
G_{i,j}=\frac{N_{i\rightarrow j}+N_{j\rightarrow i}}{N_{i}+N_{j}}\cdot E_{i,j}
\end{equation}
The binary variable $E_{i,j}$ controls the existence of an edge between $LTP_i$ and $LTP_j$. To prevent weak associations of LTPs that are not easily replaceable, we define this binary variable as:
\begin{equation}
E_{i,j}=\mathds{1}\left(\frac{\sum_k{N_{i\rightarrow k}}}{N_{i}}>\eta\right)\cdot\mathds{1}\left(\frac{\sum_k{N_{j\rightarrow k}}}{N_{j}}>\eta\right)
\end{equation}
The threshold parameter $\eta$ is set to 0.5 focusing on the elimination of LTPs via graph partitioning that are replaceable in at least $50\%$ of the training ROIs. Indeed, graph partitioning tends to preserve nodes that are not connected, which in our case would correspond to LTPs that are not easily replaced by other ones in the labeling task.

We use the Infomap algorithm \cite{infomap} to partition the similarity graph $G$.
We define the frequency of each node on $G$ as the sum of the similarity weights of connected nodes divided by twice the total weight in $G$. Then, each node is encoded with Huffman coding, where short codewords are assigned to the high-frequency nodes and long codewords are assigned to the low-frequency ones. Infomap then finds an efficient description of how information flows on the network. By detecting the partition that minimizes the description length of the network, Infomap returns a final set of sLTPs with guaranteed global optimality. 
Texture and spatial centroids $\{\overbar{FT}_{sLTP_k}, \overbar{FS}_{sLTP_k}\}$ of the sLTPs $\{ sLTP_k\}$ are then computed with Equation (\ref{FTS}) utilizing the ROIs labeled with $\{LTP_k\}$.

\subsection{Labeling of CT scans with sLTPs}\label{section_method_label}
In the test stage, scans in the whole dataset are labeled by extracting sample points and their ROIs $\{x\}$. Since it is computationally prohibitive to evaluate the textural and spatial features on every voxels within the lung masks, we only label centers of ROIs densely sampled using again SURS. Sampled ROIs with $\%emph_{-950}\leqslant1\%$ or $\%emph_{\mathrm{HMMF}}\leqslant1\%$ have their center labeled as no-emphysema class.  Remaining sampled centers get a sLTP label, via minimization of the following cost metric:
\begin{equation}\label{labelsLTP}
\chi^2 (FT_{x},\overbar{FT}_{sLTP_k}) + \lambda \cdot W \cdot ||FS_{x}-\overbar{FS}_{sLTP_k}||_2^2 
\end{equation}
Non-sampled voxels are labeled with the sLTP index of the nearest sampled center point via nearest neighbor search within the lung mask (i.e. using a Voronoi diagram). Labeling lung scans with the discovered sLTPs generates histograms of sLTPs, which are efficient lung texture signatures exploited for several tasks, as described in the evaluation sections. 

\begin{figure*}[t]
	\centering
	\includegraphics[width=.99\linewidth]{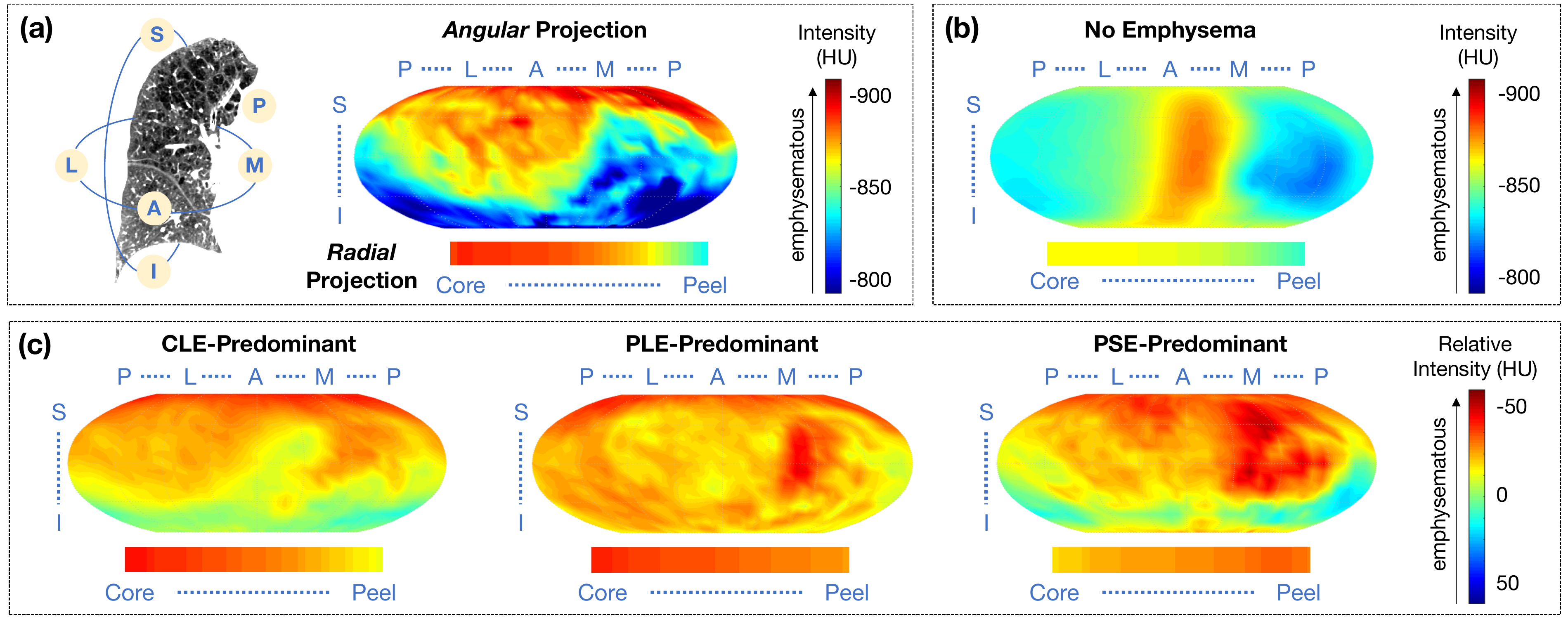}
	\caption{Population evaluation of emphysema using PDCM. (a) Illustration of superior (S), inferior (I), medial (M), lateral (L), posterior (P) and anterior (A) positions, and PDCM-based intensity projections on a sample right lung. (b) Average intensity (in HU) on PDCM-based angular  and radial projections for MESA-COPD subjects with no emphysema (N=205); (c) Average relative intensity differences, with respect to (b), on PDCM-based projections for MESA-COPD subjects with CLE-, PLE- and PSE-predominant emphysema (N= 37, 12 and 10 respectively).}
	\label{Fig:result_spatial_map}
\end{figure*}

\begin{figure*}[t]
	\centering
	\includegraphics[width=.99\linewidth]{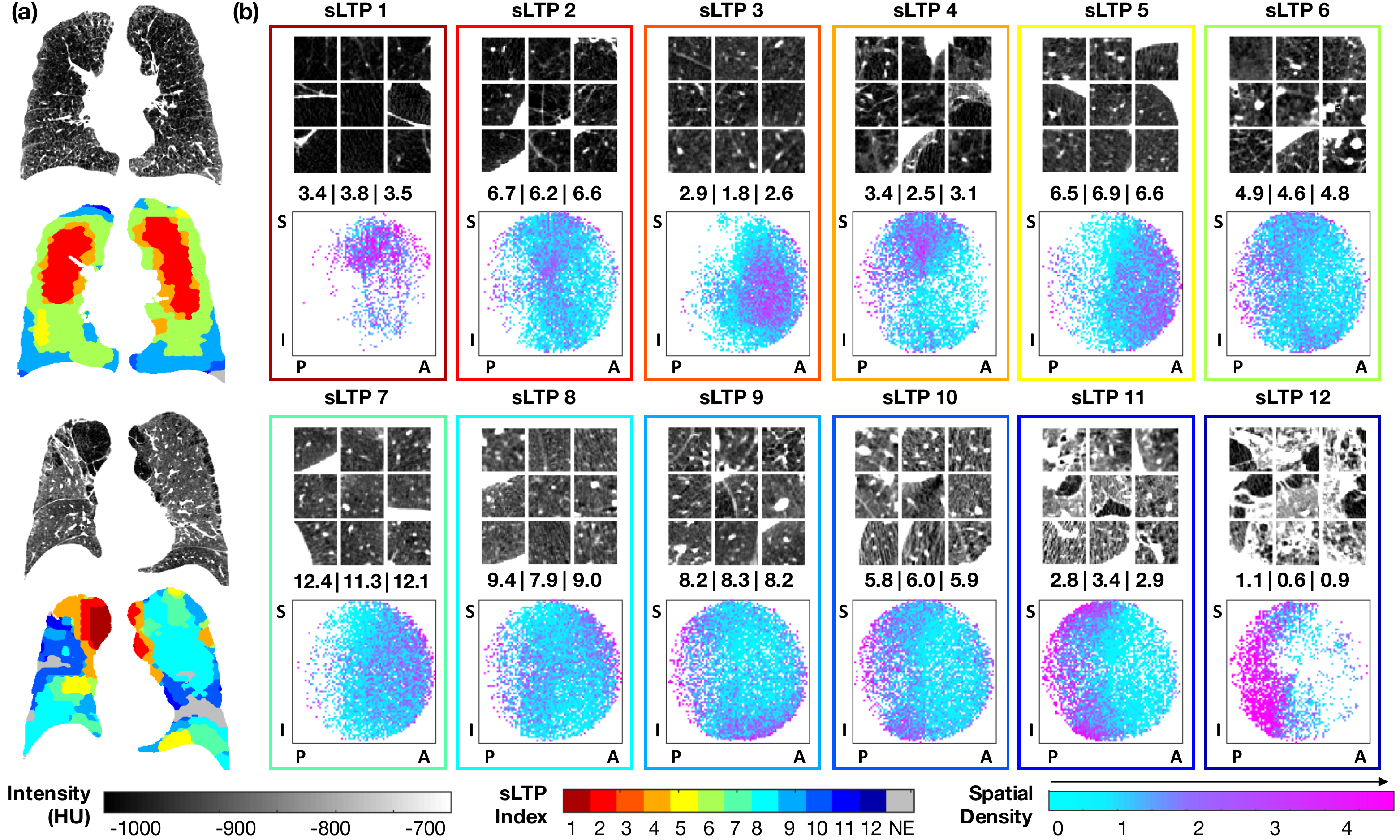}
	\caption{Qualitative illustrations of discovered sLTPs. (a) Two examples of lung scans and their sLTP labeled masks; (b) Characteristics of $\{sLTP_k\}_{k=1,..,12}$: (top) texture appearance (visualized on axial cuts from 9 random ROIs); (middle) average $\%sLTP_k$ on MESA COPD scans with $\%sLTP_k>0$ within training $|$ test $|$ all cases; (bottom) Spatial density plots of $sLTP_k$ using labeled ROIs (legend: S = superior; I = inferior; P = posterior; A = anterior positions).}
	\label{Fig:result_sltp_density}
\end{figure*}

\subsection{Spatial Density Visualization of sLTPs}\label{section_method_density}
To study the spatial distribution of sLTPs, we generate spatial visualization by scatter plotting of voxels labeled with individual sLTPs in sagittal projections, as follows. 

We first randomly sample a \textcolor{black}{initial set of ROIs} over each lung via SURS sampling. Each ROI is associated with its center point coordinates $(r,\theta,\phi)$ in the PDCMs. 
To avoid artificial higher densities on the scatter plot in regions close to the core, we adapt the number of ROIs selected per radial regions. 
The $r$ values are binned into $N_r$ intervals with midpoint values \textcolor{black}{$r_1,...,r_{N_r}$} to generate isovolumetric sub-volumes of the lung. We then define the sub-sampling ratio $\alpha_i=r_i / r_{N_r}$ (which approximates the ratio of areas in the scatter plot) and set the number of ROIs sampled per $r$ bin to $N_{\mathrm{Iso}V_i}= \alpha_i \cdot N_{\overbar{\mathrm{Iso}V}}$ where $N_{\overbar{\mathrm{Iso}V}}$ is a pre-set number of ROIs sampled in the outermost part of the lung.

All ROI centers in  \textcolor{black}{the sub-sampled set} are converted to $(x,y,z)$ Cartesian image coordinates and accumulated in a sagittal single plane, by setting $x=0$. Final density plots of sLTPs are shown in projected radial coordinates $r'=\sqrt{y^2+z^2}$ and $\phi'=atan(z/y)$. We color code each point on the sagittal projection with the following density measure: 
\begin{equation}\label{spatial_density}
Den_{sLTP_k}^{(r',\phi')}=
\frac{|\Lambda_{sLTP_k}\cap \Lambda_{(r',\phi')}|}{|\Lambda_{sLTP_k}|} \bigg/
\frac{\sum_i |\Lambda_{sLTP_i}\cap \Lambda_{(r',\phi')}|}{\sum_i |\Lambda_{sLTP_i}|}
\end{equation}
\noindent where $\Lambda_{(r',\phi')}$ denotes the set of ROIs at $(r',\phi')$ positions.
The numerator (first term) in Equation (\ref{spatial_density}) measures the probability of $sLTP_k$ at projected position $(r',\phi')$, and the denominator (second term) measures the observed overall probability of $(r',\phi')$ to host any $sLTP_i$.

\section{Experiments \& Results}\label{section_result}

\subsection{Data}\label{section_result_data}

The data used for evaluation consists of full-lung CT scans of 317 subjects. All subjects had underwent CT scanning in the MESA COPD study \cite{ben}, between 2009$-$2011. In addition, 22 out of the 317 subjects underwent CT scanning in the EMCAP study \cite{emcap}, between 2008$-$2009. 

For the MESA COPD study, all CT scans were acquired at full inspiration with either a Siemens 64-slice scanner or a GE 64-slice scanner, at 120 kVp, speed 0.5 s, and current (mA) set according to body mass index following the SPIROMICS protocol \cite{spiromics}. Images were reconstructed using B35/Standard kernels with axial pixel resolutions within the range [0.58, 0.88] mm, and 0.625 mm slice thickness. 

For the EMCAP study, scans were acquired with a Siemens 16-slice scanner, at 120 kVp, speed 0.5 s, and a current between 169 mA and 253 mA. Images were reconstructed using the B31f kernel with axial resolutions within the range [0.49, 0.87] mm, and 0.75 mm slice thickness.

Emphysema subtypes and severity have previously been assessed visually in the MESA COPD study (details available in \cite{ben}). The raters included four experienced chest radiologists from two academic medical centers. They assessed emphysema subtypes on CT scans by assigning a percentage of the lung volume affected by CLE, PLE and PSE respectively. 
Based on \cite{ben}, $N=205$ subjects do not exhibit emphysema, and are used here as the control set of no emphysema (NE) subjects. The remaining $N=112$ subjects exhibit light ($N=53$) or mild-to-severe  ($N=59$) emphysema. For these subjects, predominant emphysema subtype is defined as the subtype affecting the greatest proportion of the lungs. In the mild-to-severe cases, there are $N=37$ CLE-predominant, $N=12$ PLE-predominant, and $N=10$ PSE-predominant subjects. Overall population prevalence of emphysema in the MESA COPD cohort is 27$\%$, composed of 14$\%$ of CLE-subtype, 9$\%$ of PSE-subtype, and 4$\%$ PLE-subtype.

In addition, the following clinical characteristics are available for the scans in MESA COPD study (details in \cite{ben}): demographic factors (age, race, gender, height, weight); forced expiratory volume in 1 second (FEV1); MRC dyspnea scale measure (5-level scale); six-minute walking test (6MWT) total distance; pre (baseline) 6MWT pulse oximetry; post 6MWT pulse oximetry; reported post 6MWT fatigue; and reported post 6MWT breathlessness. We used these measures for evaluating the clinical significance of the discovered sLTP.

\subsection{Population Evaluation of Emphysema Using PDCM}\label{section_result_pdcm}

We first demonstrate the ability of our proposed PDCM lung shape mapping to study the spatial patterns of emphysema over a population of subjects (cf. Fig. \ref{Fig:result_spatial_map}). For each scan in MESA COPD study, PDCM maps of voxels inside individual lungs are generated, attributing to each voxel a coordinate $(r,\theta,\phi)$.  Voxel intensity values in PDCM maps are then averaged and visualized along two types of projections:

\begin{enumerate}[leftmargin=0.45cm,labelwidth=0.1cm]
	\item \emph{\textcolor{black}{Angular projections}}: intensity values averaged along $r$ for each pair of angular directions $(\theta,\phi)$; 
	\item \emph{\textcolor{black}{Radial projections}}: intensity values averaged over all angular directions at a subset of $N_r=60$ regular radial positions $r_1,...,r_{N_r}$. 
\end{enumerate}

An illustration of these two PDCM intensity projections on a sample lung are visualized in Fig. \ref{Fig:result_spatial_map} (a). 

Population-average PDCM angular and radial intensity projections over subjects without emphysema (NE) are displayed in Fig. \ref{Fig:result_spatial_map} (b).
\textcolor{black}{The averaged angular projection} shows a clear pattern of lower attenuations (i.e. intensity values) in the anterior versus posterior region, which agrees with the intensity gradient due to gravity-dependent regional distribution of blood flow and air \cite{gravity1963, gravity}. 
\textcolor{black}{The averaged radial projection} shows a slight gradient from core to peel regions, which is likely due to the inclusion of voxels belonging to the mediastinal and costal pleura inside the lung mask.

Population-average PDCM intensity projections over subjects with CLE-, PLE-, and PSE-predominant emphysema subtypes are visualized in Fig. \ref{Fig:result_spatial_map} (c). To highlight differences with respect to the control set, \textcolor{black}{we display relative values after subtraction of the values from the corresponding NE average projection in Fig. \ref{Fig:result_spatial_map} (b)}. Color coding represents relative intensity differences with more emphysema (more negative attenuation values) corresponding to the red color.

We can see on the relative \emph{angular} PDCM intensity projections that regions of normal attenuation (green to blue) are absent for PLE-predominant subjects, whereas CLE- and PSE-predominant subjects appear to have emphysema regions (red) concentrated in the superior part. The average relative \emph{radial} PDCM intensity projections on emphysema subjects show systematic higher attenuation values, with more emphysema in the core part for CLE-predominant subjects and more emphysema in the peel part for PSE-predominant subjects.

\begin{table}
\textcolor{black}{\caption{Parameter Setting for sLTP Learning.}}.
\centering
\small{\color{black}
\begin{tabular}{L{2.75cm}L{5.0cm}}
	\hline
	Parameters & Setting \\
	\hline
	\noalign{\medskip}
	ROI size   										& = 25 mm$^3$, to approximate the size of secondary pulmonary lobules \\
	\noalign{\medskip}
	$\beta_1$: random shift					 	    & $\in [0,25]$ mm  \\
	(for ROI sampling) 	    						& \\
	\noalign{\medskip}
	$\beta_2$: sample density				 	    & = 3 samples per stack  \\
	(for ROI sampling) 								&  \\
	\noalign{\medskip}  	
	\# of textons: 								    & = 40, targeting 10 textons per \\
	(for texture feature)							& standard emphysema subtype and  \\
													& normal tissue class, according to \cite{texton_miccai2010} \\
	\noalign{\medskip} 
	Texton size										& 3$\times$3$\times$3 pixels, according to \cite{ ltp_mcv16} \\
	\noalign{\medskip} 
	\# of lung sub-regions							& = 36, according to binning of $(r,\theta,\phi)$ \\
	(for spatial feature)							& in Section \ref{section_spatial_feature}.\\
	\noalign{\medskip}
	$N_{LTP}$: \# of LTPs in initial set   			& = 100, as suggested in \cite{ltp_mcv16}), 
													  for sufficient diversity of the patterns 
													  and being able to discover rare emphysema types\\
	\noalign{\smallskip}
	\hline
\end{tabular}}
\label{Table:parameter}
\end{table}
	
\subsection{Qualitative Evaluation of Discovered sLTPs}\label{section_result_visual}
For the discovery of sLTPs, 3/4 of the total scans in MESA COPD study (N=238) were used for training, using random stratified sampling without replacement, while the other scans (N=79) were used for testing. \textcolor{black}{We summarize the setting of pre-defined parameters for the sLTP learning in TABLE I.} In addition, spatial regularization weight $\lambda$ is set via empirical tuning using Eq. (\ref{tune-lambda}). Based on the relative texture homogeneity loss measure $\Delta SSW_T$, we chose $L_T=1\%$ which corresponds to $\lambda=1.52$, above which $\Delta SSW_T$ increases drastically.

A total of 12 sLTPs were discovered using the full training set, and were used to label both the training and test scans in emphysema-like lung. Each sLTP was detected (i.e. $\%sLTP_k>0$) in at least 5\% of scans both in training and test sets. In Fig. \ref{Fig:result_sltp_density}, we illustrate in (a) the sLTP labeling of two sample CT scans; and in (b) the characteristics of each sLTP via visual illustrations of labeled patches, average occurrence in MESA COPD scans, and spatial distribution of their occurrence within the lungs. For the patch illustrations, 9 samples were randomly selected from all available labeled ROIs. For the average occurrence, we averaged $\%sLTP_k$ values over scans with $\%sLTP_k>0$. For the spatial distributions, we generated spatial scatter plots of sLTP locations from labeled ROIs, following the method described in \ref{section_method_density}, with $N_{\overbar{\mathrm{Iso}V}}=5,000$, and $N_r=60$. 

We can observe that patches belonging to an individual sLTP appear to be textually homogeneous. sLTP 1 and 4 show clear spatial accumulation in superior (apical) regions, sLTP 3, 5 and 7 in anterior regions, and sLTP 10, 11 and 12 in posterior regions. All sLTPs returned similar occurrences in training and test sets. Some sLTPs are rare, such as sLTP 12 which covers $\sim$1\% of the lungs when present, but is still found in 24 scans over the whole MESA COPD cohort. 

\subsection{Reproducibility of sLTPs}\label{section_result_repro}

\medskip
\subsubsection{Reproducibility of sLTP labeling versus training sets}
To test the reproducibility of sLTPs learning, we first compare the $N_{\mathrm{sLTP}}=12$ sLTPs $\{sLTP_k\}$ generated with the full set of training scans, to $N_{\mathrm{set}}=4$ sLTPs sets $\{sLTP_k^c\}_{(c=1,2,3,4)}$ using subsets of training data by randomly eliminating 25\% of the training scans. Reproducibility of sLTPs is evaluated on the ROI labeling task, by computing the average overlap of labeled test ROIs with the following metric:
\begin{equation}\label{Rln}
R_{\mathrm{ln}}=\frac{1}{N_{\mathrm{set}}\cdot N_{\mathrm{sLTP}}}
\sum_{c=1}^{N_{\mathrm{set}}}\sum_{k=1}^{N_{\mathrm{sLTP}}}
\frac{|\Lambda_{sLTP_k} \cap \Lambda_{\pi(sLTP_k^c)}|}{|\Lambda_{sLTP_k}|}
\end{equation}
where $\Lambda_{sLTP_k}$ denotes the set of ROIs labeled with $sLTP_k$, and $\pi()$ denotes the permutation operator on the $\{sLTP_k^c\}$  determined by the Hungarian method \cite{resampling} for optimal matching between sets $\{sLTP_k\}$ and $\{sLTP_k^c\}$. 

Compared with the $N_{\mathrm{sLTP}}=12$ sLTPs learned on the full training set, we discovered $N^c_{\mathrm{sLTP}}=$ 12, 12, 13, and 13 sLTPs on training subsets. We obtain an overall labeling reproducibility measure of $R_{\mathrm{ln}}=$ 0.91 which corresponds to a high reproducibility level.

\textcolor{black}{We then further compute the reproducibility measure, denoted as $R_{\mathrm{ln}}'$, among training subsets. The metric is similar to Equation \ref{Rln}, replacing $\{sLTP_k\}$ and $\{sLTP_k^c\}$ with sLTPs $\{sLTP_k^{c1}\}$ and $\{sLTP_k^{c2}\}$ ($c1 \neq c2$) learned on different training subsets. We obtain an overall labeling reproducibility measure of $R_{\mathrm{ln}}'=$ 0.85 (standard deviation = 0.07)}

\textcolor{black}{To evaluate the contribution of spatial features in sLTP learning, we further generate sets of lung texture patterns using only texture features (i.e. using initial LTPs without spatial augmentation in Section \ref{section_method_ltp}, and setting $\lambda=0$ for the replacement test in Section \ref{section_method_sltp}). We discovered 11 patterns using the full training set, and 11, 11, 12 and 12 patterns on training subsets. The reproducibility measures $R_{\mathrm{ln}}$ and $R_{\mathrm{ln}}'$ equal to 0.84 and 0.78 (standard deviation = 0.12), are lower than the ones obtained using the proposed sLTP learning, hence confirming the benefit of adding spatial features.} 

\medskip
\subsubsection{Reproducibility of sLTP labeling versus ROI sampling}
\begin{figure}
	\centering
	\includegraphics[width=.875\linewidth]{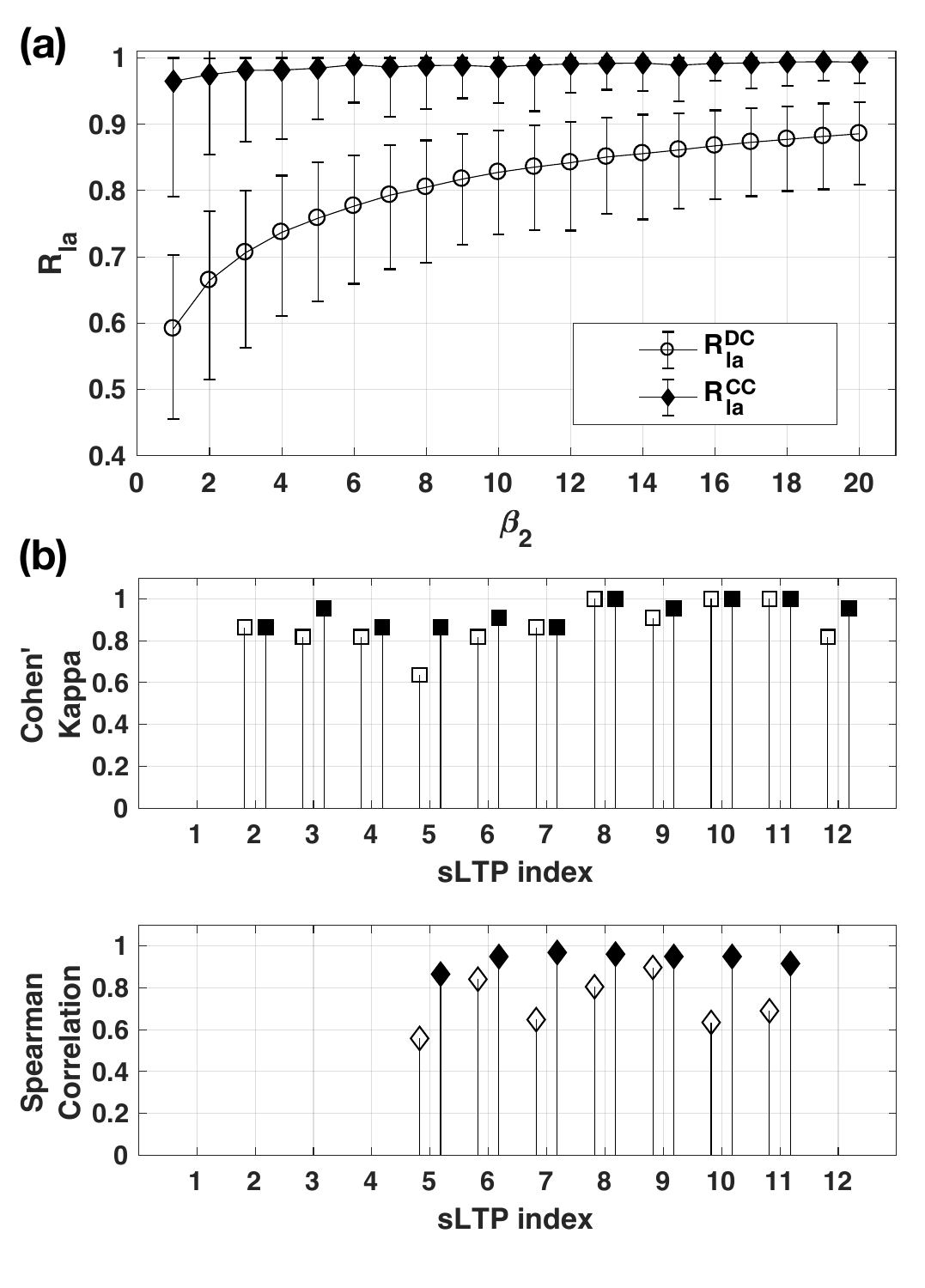}
	\caption{Results of sLTP reproducibility measures. (a) Reproducibility measures $R_{\mathrm{la}}$ versus ROI sampling parameter $\beta_2$; (b) Reproducibility of sLTPs labeling across scanners (from EMCAP and MESA COPD studies) measured with Cohen's Kappa coefficients of $sLTP_k$ presence and Spearman correlation coefficients of $\%sLTP_k$ values (white = without and black = with intensity histogram mapping).}
	\label{Fig:result_repro}
\end{figure}

As detailed in Section \ref{section_method_label}, sLTP labeling is based on a subset of voxels setting ROI positions, using SURS-based sampling strategy, which is controlled with the parameter $\beta_2$ (number of samples per stack). The selected ROIs have an influence on the final outline of the label map, which is hopefully minor if ROIs are sampled densely enough  and if sLTPs are generic enough. In this experiment, we test this hypothesis by generating two different sets of ROIs on test scans using two different random seedings, and measure the reproducibility of the generated label masks using the $\{sLTP_k\}$ discovered on the full training set, while varying the $\beta_2$ parameter. We measure labeling reproducibility using the two sets of ROIs with the following metrics:
\begin{itemize}[leftmargin=0.4cm,itemindent=0cm,labelwidth=\itemindent,labelsep=0.2cm,align=left]
\item $R_{\mathrm{la}}^{DC}(sLTP_k,\beta_2)$ = average of Dice coefficients of label masks of $sLTP_k$ over all test scans; 
\item $R_{\mathrm{la}}^{CC}(sLTP_k,\beta_2)$ = Spearman correlation coefficients of $\%sLTP_k$ values within the lungs over all test scans. 
\end{itemize}
We illustrate in Fig. \ref{Fig:result_repro} (a), the average, max and min values of $R_{\mathrm{la}}^*$ measures over all $\{sLTP_k\}$, for $\beta_2 \in [1,\ 20]$. Both reproducibility measures increase with $\beta_2$ in an exponential manner. We obtain an average $R_{\mathrm{la}}^{DC}>0.8$ when $\beta_2>10$, corresponding to sampling less than $0.05\%$ points in each stack. We obtain an average $R_{\mathrm{la}}^{CC}>0.9$ when $\beta_2>5$. Minimum $R_{\mathrm{la}}$ values always occur for sLTP 12, which is the rarest sLTP, as reported in Section \ref{section_result_visual}. 

\medskip
\subsubsection{Reproducibility of sLTP labeling versus scanner type}
The 22 subjects from MESA COPD previously scanned within the EMCAP study, underwent different generations of CT scanners. \textcolor{black}{This subset of population is relatively normal. The average time lapse between EMCAP and MESA COPD scans is 14-months. The mean of $\%emph_{-950}$, calibrated for outside air values, is 0.7\% (min $<$ 0.1\%, max = 3.9\%) in EMCAP, and 2.6\% (min = 0.3\%, max = 9.5\%) in MESA COPD, corresponding to an average increase of $\%emph_{-950}$ equal to 1.9\%.
Therefore, we use this subset of scans to evaluate the reproducibility of sLTP labeling versus scanner types.}

We used the 12 sLTPs discovered on the full MESA COPD training set. Because of differences in scanner generations (axial CT in EMCAP versus spiral CT in MESA COPD) and radiation dose settings, intensity calibration was required, implemented in two steps: 1) equalizing the outside air mean intensity value (according to \cite{hmmf_tmi14}); 2) histogram mapping of normal lung parenchyma identified with the HMMF-based emphysema masks. The sLTPs 2 to 12 were found to be present in both datasets, but sLTPs $\{2, 3, 4, 12\}$ occur in less than 6 pairs of scans. 
We report in Fig. \ref{Fig:result_repro} (b) the Cohen's Kappa coefficients of $sLTP_k$ presence for sLTPs 2-12, and the Spearman correlation coefficients of $\%sLTP_k$ for the frequent sLTPs only (sLTPs 5 to 11).
The Cohen's Kappa coefficients and Spearman correlations are all above 0.8, which confirms robust sLTP presence and percentage labeling on the 22 subjects scanned on different scanner types in two studies.

\begin{figure}
	\centering
	\includegraphics[width=.975\linewidth]{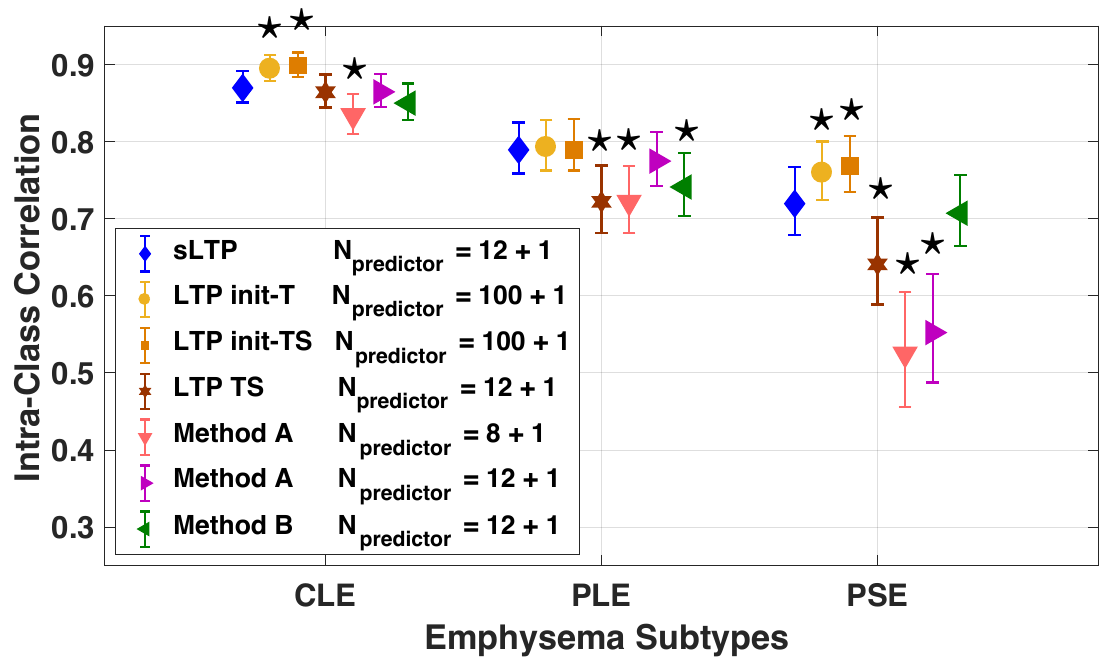}
	\caption{Intraclass correlation (ICC) and 95\% confidence interval between predicted standard emphysema subtype scores and ground-truth. Differences with sLTP-based values are marked as $\star$ when significant ($p<0.05$).}
	\label{Fig:result_icc}
\end{figure}

\subsection{sLTPs' Ability to Encode Standard Emphysema Subtypes}\label{section_result_icc}
When generating unsupervised lung texture patterns \textcolor{black}{(either sLTPs in this work or earlier generations of LTPs in previous work)}, we expect them to be finer-grained than the three standard emphysema subtypes used in \cite{ben}, while still capable to encode them, hence linking unsupervised image-based emphysema subtyping with clinical prior knowledge. 

The \textcolor{black}{LTPs (or sLTPs)} can be interpreted as either pure or a mixture of the three standard subtypes. We hereby evaluate the ability of the generated \textcolor{black}{LTPs (sLTPs)} to predict the overall extent of standard emphysema subtypes. To do this, we generate, for each scan and per lung, two signature vectors: 1) a LTP signature histogram composed of the percentage of non-emphysema class (obtained as in Section \ref{section_method_label}) and the percentages of individual \textcolor{black}{LTP (sLTP)} in the emphysema-like lung. This normalized histogram is called the LTP predictor signature and is of size $N_{\mathrm{predictor}}=N_{\textcolor{black}{LTP}}+1$; 2) a ground-truth signature composed of the percentage of non-emphysema and the three standard emphysema subtypes, as visually evaluated in \cite{ben}. 
A constrained multivariate regression model is used on labeled training scans to learn regression coefficients between the LTP and ground-truth signatures, using the following optimization:
\begin{equation}\label{regress}
\mathrm{argmin}_A\|XA-Y\|^2_2 \ \ \mathrm{s.t.}\ 0<A_{k,i}<1\ \mathrm{and}\ \sum\nolimits_{i}A_{k,i}=1
\end{equation}
where $X_{N_{\mathrm{scan}}\times N_{\mathrm{predictor}}}$ is composed of all training  LTP signatures in $N_{\mathrm{scan}}$ training scans, and $Y_{N_{\mathrm{scan}}\times4}$ contains the ground-truth signatures. $A_{N_{\mathrm{predictor}}\times4}$ is the matrix of regression coefficients $\{A_{k,i}\}$, which measure the probability of a voxel labeled as a certain predictor belonging to one of the ground-truth classes, and are therefore constrained to be in the range of $[0,1]$. Optimization of regression was solved using the CVX toolbox (\url{http://cvxr.com/cvx}).

\begin{figure}
	\centering
	\includegraphics[width=.975\linewidth]{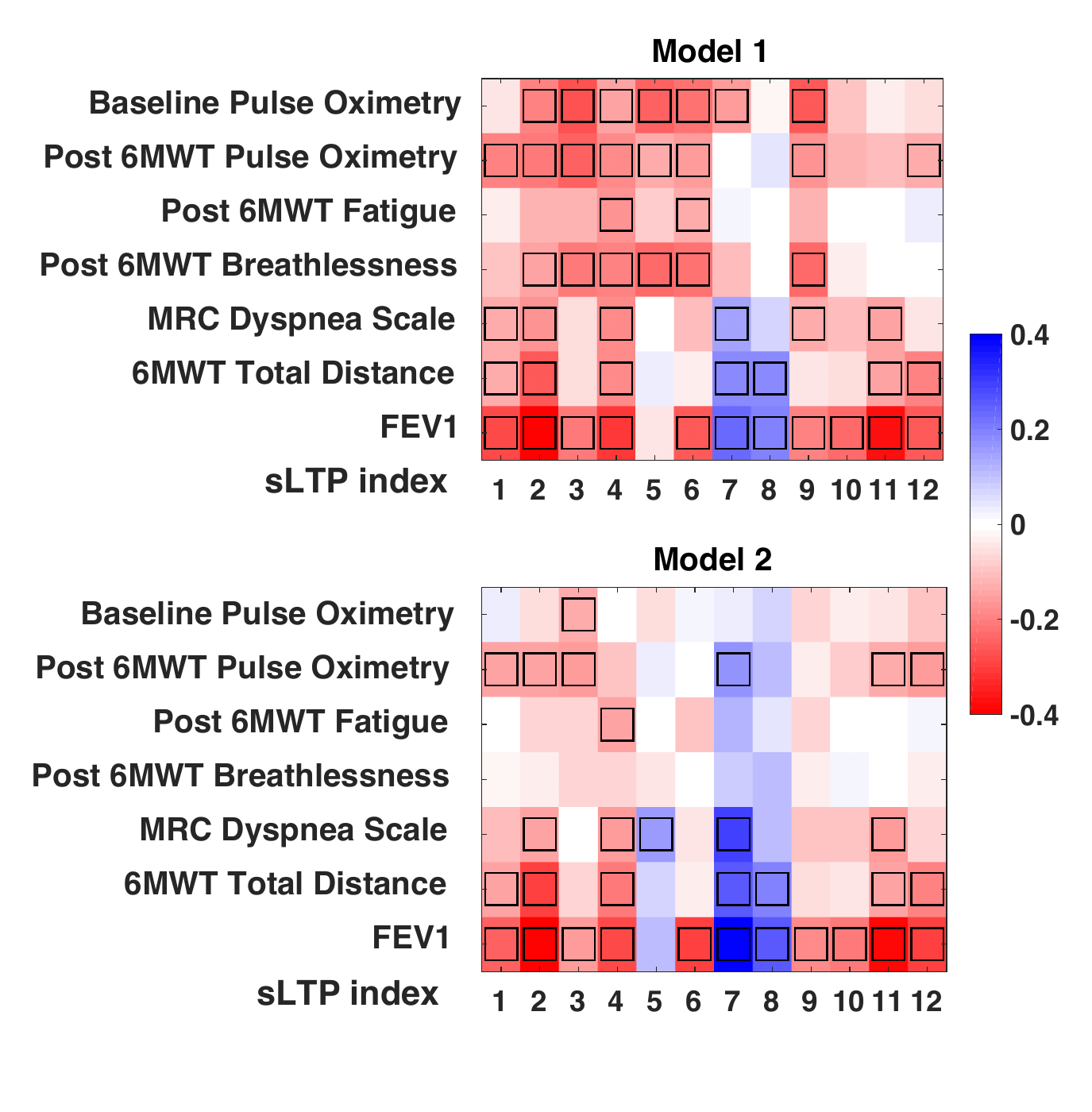}
	\caption{Partial correlations between $\%sLTP_k$ and clinical measures after adjusting for demographical factors (Model 1), and adjusting for demographical factors and $\%emph_{-950}$  (Model 2). Black-boxes indicate statistically significant values ($p<0.05$).}
	\label{Fig:result_clinc}
\end{figure}

Quality of prediction is measured with the intraclass correlation (ICC) between predicted and ground-truth exploiting \textcolor{black}{the full MESA COPD dataset}. We use a 4-fold cross validation (3/4 label masks used for training the regression and 1/4 used for testing and measuring prediction quality). Significance of differences in ICC values was assessed using Fisher’s r-to-z transformation and a two-tailed test of the resulting z-scores.

In Fig. \ref{Fig:result_icc}, we compare prediction quality with 7 sets of \emph{emphysema-specific} LTPs (re)trained on the same set of emphysematous ROIs: 1) the 12 sLTPs learned in this study; 2-3) the initial set of 100 LTPs generated in this study before (denoted as LTP init-T) and after (denoted as LTP init-TS) spatial augmentation; \textcolor{black}{4) LTPs generated by one-stage clustering (denoted as LTP TS) of the proposed texture and spatial features, by setting $N_{LTP}=12$ directly (this is to test the contribution of the proposed two-stage learning in Section \ref{section_method_ltp});} 5-6) LTPs re-generated using Method A \cite{ltp_isbi15}, discovered via graph partitioning of 100 candidates based on local spatial co-occurrence and with $N_{LTP}=8$ as in \cite{ltp_isbi15} or 12; 7) LTPs re-generated using Method B \cite{ltp_mcv16}, discovered via merging 100 candidates based on texture similarity and local spatial co-occurrence, and setting $N_{LTP}=12$ for the iterative merging.

\textcolor{black}{Fig. \ref{Fig:result_icc} shows that the two sets of 100 LTP models achieve overall best prediction accuracy, and that the newly discovered 12 sLTPs have the best performance among the 5 small LTP sets. Difference of ICC values between the sLTPs and the 100 LTP models was not significant for PLE emphysema subtype.}

\subsection{Clinical Associations of sLTPs}\label{section_result_clic} 
To evaluate clinical association of sLTPs, we first compute Spearman's partial correlations between $\%sLTP_k$ within both lungs and the seven clinical characteristics listed in \ref{section_result_data}, \textcolor{black}{on the full MESA COPD dataset}, using two models: Model 1 adjusted for demographical factors (age, race, gender, height and weight), and Model 2 further adjusted for $\%emph_{-950}$. The results are reported in Fig. \ref{Fig:result_clinc}. Correlation values for MRC dyspnea scale, post 6MWT breathlessness and post 6MWT fatigue are flipped in the figure so that more negative correlation values always correspond to more severe symptoms. 

Overall, we obtained 47 and 31 significant correlations with Models 1 and 2. The sLTPs 7 and 8 are associated with healthier subjects (positive correlations), while the other sLTPs correlate with symptoms (negative correlations). In Model 1, all clinical variables show significant correlations with 2 to 11 sLTPs. \textcolor{black}{While applying similar setting to the standard subtypes, only CLE and PLE show significant associations with MRC dyspnea scale and 6MWT total distance, and only CLE show significant associations with FEV1, as reported for the same population in \cite{ben}.}

Model 2 looses significant correlations for post 6MWT breathlessness, but preserves all, or almost all, significant correlations for FEV1, 6MWT total distance, dyspnea and post-6MWT oximetry.

\textcolor{black}{We then further adjust for FEV1 in Model 2. In this rigorous setting, sLTP 3 remains significantly correlated with pre- and post-6MWT oximetry; sLTP 2, 4 and 7 remain significantly correlated with 6MWT total distance, and sLTP 7 remains significantly correlated with MRC dyspnea scale. While applying similar setting to the standard subtypes, only CLE and PLE show significant associations with 6MWT total distance \cite{ben}.}

\section{Discussion $\&$ Conclusion}

In this work, we propose a novel unsupervised learning framework for discovering \emph{emphysema-specific} lung texture patterns on the MESA COPD cohort of CT scans. The proposed method incorporates spatio-textural features via an original cost metric combining \raisebox{1pt}{$\chi^2$}-$\ell^2$ constraints,  along with data-driven parameter tuning, and Infomap graph partitioning. 

Our methodological framework includes the introduction of a standardized spatial mapping of the lung shape utilizing Poisson distance map and conformal mapping to uniquely encode 3D voxel positions and enable comparison of CT scans \textcolor{black}{without registration being required further than orientation alignment}. 
Our lung shape spatial mapping PDCM enables straightforward population-wide study of emphysema spatial patterns. \textcolor{black}{By visualizing relative \emph{angular} PDCM intensity projections on CLE-, PLE- and PSE-predominant subjects, we can see that regions of normal attenuation are absent for PLE-predominant subjects, which agrees with the definition of PLE (diffused emphysema subtype). CLE- and PSE-predominant subjects appear to have emphysema regions concentrated in the superior part. This agrees with the observation made in \cite{ben} on the same dataset that CLE and PSE severity was greater in upper versus lower lung zones, whereas severity of PLE did not vary by lung zone. By visualizing relative \emph{radial} PDCM intensity projections, we can see that emphysema subjects show systematic higher attenuation values than subjects without emphysema, as expected. CLE-predominant subjects have more emphysema in the core part, whereas PSE-predominant subjects have more emphysema in the peel part. This agrees with the definitions of CLE and PSE.
As a standardized tool, the proposed spatial mapping PDCM is not tied to emphysema pattern, and our future work will exploit such spatial mapping to study other pulmonary diseases.}

\textcolor{black}{With the proposed method, we discovered 12 spatially-informed lung texture patterns (sLTPs) on the MESA COPD cohort. 
Qualitative visualization show that the discovered sLTPs appear to be textually homogeneous with different spatial prevalence. Since we jointly enforce spatial prevalence and textural homogeneity, each sLTP can have spatial ``outliers'' that are texturally favored. Extensive evaluations show that the discovered sLTPs are reproducible with respect to training sets, sampling of ROI for labeling, and certain scanner changes. The proposed incorporation of spatial and texture features obtains higher learning reproducibility compared to using texture features only, confirming the benefit of spatial regularization. 
The number of discovered sLTPs varies slightly between training subsets. This can be caused by a large change in the proportion of rare LTPs within the our subsets, which modifies the weights in the Infomap similarity graph. A larger dataset with more diseased cases might be beneficial to solve this issue.}

\textcolor{black}{The sLTPs are able to encode the three standard emphysema subtypes, and thus link unsupervised discovery with clinical prior knowledge.
Prediction quality is better than previous models, and close to the optimal level reached with 100 \emph{emphysema-specific} LTPs.
While intra-cluster LTP homogeneity increases with the number of LTPs, hence  leading to higher prediction performance, working with 100 LTPs leads to redundancy between subtypes which is detrimental when studying associations of individual LTPs with clinical measures. One-stage clustering leads to significantly lower prediction power for PLE and PSE subtypes, compared to sLTPs, which demonstrate the benefit of the proposed two-stage learning.}

Significant correlations with physiological symptoms were found for several measures. Training our discovery of \emph{emphysema-specific} sLTPs on ROIs with $\%emph>1$ aimed to enable discovery of early emphysema stages. Our correlation results suggest that sLTPs 7 and 8 are good candidates for early emphysema characterization, not yet associated with physiological symptoms.
\textcolor{black}{Significant correlation results after adjusting for$\%emph_{-950}$ indicate that our sLTPs provide clinically-relevant and complementary information to the commonly used $\%emph_{-950}$ measure. In the rigorous setting after adjusting for FEV1, there are still sLTPs showing significant correlations with MRC dyspnea scale, 6MWT total distance, pre- and post-6MWT oximetry. While for the standard emphysema subtypes, only CLE and PLE remain significantly associated with 6MWT total distance.}

Progression patterns of the sLTPs will be investigated in the future, via sLTP labeling of longitudinal CT scans (with large time lapse). 
The sLTP histograms extracted in this study provide texture signatures that can be used to characterize and group CT scans. Patient grouping was found beneficial to study physiological indicators of COPD in \cite{ltp_mit16}, and will be considered in our future study.   
Further development is possible to improve the generation of image-based sLTPs with demographic and population-wide information, which would likely reveal population-specific and population-invariant patterns, but requiring a larger and more diseased cohort for training. 

\section{Acknowledgments}

The authors sincerely thank the investigators, the staff, and the participants of the MESA study (\url{http://www.mesa-nhlbi.org}) for their contributions to this valuable dataset. The authors would also like to thank Dr. Jingkuan Song for technical advice and valuable comments.

\bibliographystyle{IEEEtran}
\bibliography{sltp_extension_2020final_ARXIV}

\end{document}